\documentclass[runningheads,a4paper]{llncs}
\usepackage{float}
\usepackage{cite}
\usepackage{graphicx}
\usepackage{filecontents}
\begin{document}

\title{Visual  Analytics  of  Movement Pattern Based on Time-Spatial Data: A Neural Net Approach}
\titlerunning{Visual  Analytics of Movement Pattern}  % abbreviated title (for running head)
\author{Zhenghao Chen\inst{1}\and Jialong Zhou\inst{2}\and Xiuying Wang\inst{1}}
\authorrunning{Z.Chen et al.} % abbreviated author list (for running head)

%
\institute{School of Information Technologies,
University of Sydney, NSW 2006, Australia\\
\email{\{zhenghao.chen, xiu.wang\}@sydney.edu.au}\\
\and Data 61, CSIRO, NSW 2006, Australia\\
\email{\{jialong.zhou\}@data61.csiroau}}

\maketitle              % typeset the title of the contribution

\begin{abstract}
Time-Spatial data plays a crucial role for different fields such as traffic management. These data can be collected via devices such as surveillance sensors or tracking systems. However, how to efficiently analyze and visualize these data to capture essential embedded pattern information is becoming a big challenge today. Classic visualization approaches focus on revealing 2D and 3D spatial  information and modeling statistical test. Those methods would easily fail when data become massive. Recent attempts concern on how to simply cluster data and perform prediction with time-oriented information. However, those approaches could still be further enhanced as they also have limitations for handling massive clusters and labels. In this paper, we propose a visualization methodology for mobility data using artificial neural net techniques. This method aggregates three main parts that are Back-end Data Model, Neural Net Algorithm including clustering method Self-Organizing Map (SOM) and prediction approach Recurrent Neural Net (RNN) for extracting the features and lastly a solid front-end that displays the results to users with an interactive system. SOM is able to cluster the visiting patterns and detect the abnormal pattern. RNN can perform the prediction for time series analysis using its dynamic architecture. Furthermore, an interactive system will enable user to interpret the result with graphics, animation and 3D model for a close-loop feedback. This method can be particularly applied in two tasks that Commercial-based Promotion and abnormal traffic patterns detection. 
\keywords{Visual Analytics, Deep Learning, Prediction, Clustering}
\end{abstract}
\section{Introduction}
\begin{figure}[!h]
\centering
\includegraphics[width=0.9\linewidth]{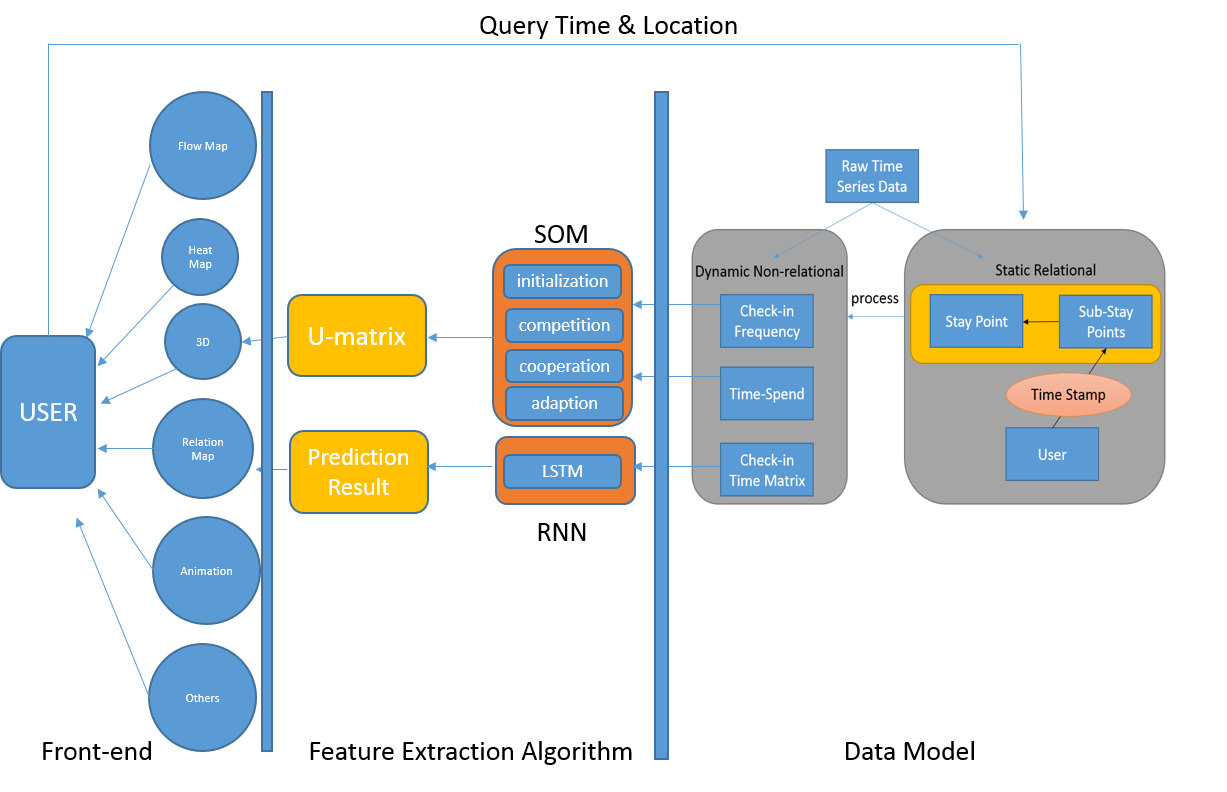}
\caption{\label{fig:ff} Framework of Our Method}
\end{figure}

Time-spatial data consists of spatial and temporal contextual information,  which is a complicated type of data set. Nowadays, with the development of technique, it is becoming more and more convenient to collect such data with detailed day and time information than it was ever before  \cite{cnn}. Taking mobility data as an example which represents the objects movement in real time with recorded the geographical information   \cite{von2016mobilitygraphs}, this type of data can be collected using Global Positioning System (GPS), sensors or tracking systems. The information embed from those data such as behavior pattern can be also implied in urban development, traffic transportation, and big event management. Therefore the understanding of this particular part of data is essential.
To obtain meaningful results from those data, this paper proposes a methodology that utilizes neural network together with other visualization approaches. The framework of our method is shown as Figure\ref{fig:ff}. There are three majority parts. Firstly, this approach utilizes the relational database combined with dynamic non-relational database as back-end. Temporal database will be generated once user requests from the relational database. The second part is features extraction algorithm. We propose two approaches prediction and clustering. For clustering, SOM will be applied to cluster visiting pattern in order to group the visitors and find the abnormal groups that can be further suspected as crime or others. Moreover, the inspections will be put on such abnormal clusters. Regarding prediction, LSTM network is utilized to predict upcoming locations that travelers would like to go in next time-stamp. Eventually, the extracted information such as clustering and prediction result will be displayed using graphics and multi-media techniques including 3D time-spatial cube, 2D heat map, relational map and animation to users. Such displays will help users understand behavior pattern of moving objects better. 

Using our approach, we participated in the public challenges that Visual Analytics Science and Technology Challenge 2017 (VAST 2017) as well as assist a mobile company in China for their promotion activity.  In terms of commercial based visualization such as mobile users promotion \cite{cnn}, we used this method to obtain user behavior pattern. Based on those information, we can provide our suggestions to the company manger. On the other hand, we take traffic visualization and prediction in VAST 2017 to research how movement of vehicles affect wildlife via understanding their pattern by conducting the pipeline of visual analytics. Consequently, we provide suggestions to a Chinese mobile company for user promotion and successfully predict the affect of vehicles to wildlife in a nature preserve. Overall contributions can be summarized as:

\begin{itemize}
\item We set up a time-spatial data visual analytics pipeline to understand movement behavior.

\item We modify and utilize neural net approach to propose a clustering and prediction methodology .

\item We use this approach to resolve two data analytics problems including commercial promotion as well as pattern detection and prediction.

\end{itemize}
\section{Related Work}
The previous work are main concerned on two parts that clustering and aggregation as well as feature learning.\cite{zheng2009mining} \cite{andrienko2013visual}.

\subsection{Clustering and Aggregation}
Increasing amount of data collection, analysis of trajectories is becoming even more massive. Therefore, single mobility data analysis and visualization become un-meaningful. The state of the art research focus on aggregation, clustering and grouping in order to simplify time-spatial context information via reducing the size of data representation  \cite{von2016mobilitygraphs}. With those advance techniques, the visualization of time-series data becomes much more elegant than before. The biggest challenge for those research is to generate an efficient algorithm for aggregation, grouping and clustering.

\subsection{Feature Learning}
The big breakthrough was conducted with machine learning. Classic clustering methods such as Gaussian Mixture Model (GMM) with Expectation Maximization (EM) and K-means serves as effective tool. However, clustering raw trajectory with multiple-dimension is not sufficient for larger time-spatial data set  \cite{von2016mobilitygraphs}. Recently, the attempts of using Neural Network Self-Organizing Map (SOM)  \cite{kohonen1998self} to cluster the trajectory data is done in the 2012  \cite{shukla2012self}. That research combines temporal-context information and other associated data such as social-economics background of users. Using such data, researchers tried to visualize the pattern of different groups of visitors and their potential to commit crime. In 2015, the T.Landersberger used spatial-simplification algorithm  \cite{von2016mobilitygraphs} to dynamically reduce the flow information after aggregation and clustering which is the latest approach to achieve the aggregation task so far. State-of-art method deep learning enables the prediction of time-spatial for next timestamp. Researchers start looking at how to predict the upcoming location of movement  \cite{liu2016predicting} and generate an inference or generative model. Recurrent Neural Net (RNN)  \cite{medsker2001recurrent} is one of the popular deep learning models which makes it easier to deal with vectors where each dimension has strong dependency from previous one.\cite{liu2016predicting}. 

\section{A Dynamic to Static Data Model}
There are two main types of data. For one, we have relational database as static database. That is all the data will be stored in the relational database model using relationship $Object - [Activity] -> StayPoint$. Also, both stay points and objects are stored using tree data structure. Specifically, stay point will have sub-stay points as generalization. Moreover, non-relational dynamic database will be generated from static database once user requests. Those data will consist of check-in frequency, time-spend, squence and Time-Oriented matrix.

\textbf{Visiting-Frequency Matrix} It is to indicate how many times a visitor visit in every location, we use visiting frequency matrix. 

\textbf{Time-Spent Matrix} Simply using frequency count might lost a lot of information of visitor behavior information. Hence, we also need to consider the period that visitor spend in certain attraction.

\textbf{Sequence Vector} Sequence vector records the location that a traveler visits in certain period without considering time information. Such vector will simply track places a traveler check in using a one-dimensional vector.

\textbf{Time-Oriented Matrix} Time-oriented matrix have two dimensions for representing user-location-traveling-stay information. Specifically, X-dimension represents the users list and Y-Dimension represents the time that is divided into certain time duration. The value of matrix will be the discrete identification value of each location that a traveler stays in at certain time stamp. 

\section{Neural Net Based Feature Extraction Algorithms}
 This section will introduce the Neural Net Based Feature Extraction Algorithms including SOM-based clustering and RNN-based Prediction, the algorithm is inspired from our previous work\cite{cnn}. We use these two different algorithms for generating predicted flow map and clustering Unified Matrix(U-matrix). Our main contribution is concerned on the post-training visualization rather than modifying the neural net.

\subsection{Self-organizing Map based Clustering}
An unsupervised approach Self-organizing Map would be utilized here. Furthermore, most of previous work tried to cluster the movement trajectory directly. However, such clustering become very difficult for clustering large and irregular trajectories. Different from those, in this approach, I cluster moving people based on their preferences of locations (time spent and frequency matrix). This clustering method helps to find people who have the high potential to be in the same groups once they have similar or even the same interests. Based on those, we can further infer what category of visitors they are. For instance, a group of family with children moving in a park would probably have higher check-in frequency on kiddie rides than a group of teenagers.

Essentially, there are two parts for this work. For one, we use SOM to train and render a feature map after clustering. For another, after clustering result is rendered, we use associated visual analytics approach to recognize different patterns from the rendered Unified Distance Matrix from SOM. There are couple of essential steps including \textbf{Initialization}, \textbf{Competition}, \textbf{Cooperation} \textbf{Adaption} \textbf{Group Aggregation} and \textbf{Visiting Pattern Analysis}. The detail of those parts will be explained in the following sections.

In addition, associated visual analytics is crucial in this approach. Using those approaches to visualize the patterns such as visitor preferences on different locations will be useful to understand movement behaviors. Besides, other features including communication, time spent on a site, preference categories are also considered as essential features for clustering.

\subsection{Deep Learning based Prediction}
As we process the time-oriented matrix as dynamic data which indicates where visitor stays at each timestamp. We will use this matrix as training data to forecast the travel flow of visitors. By using this, we are not going to lose any time information which results in better performance than using pure sequence matrix \cite{liu2016predicting}. That is because duration that individual stay in a location is an essential factor of behavior choice  \cite{zhang2016deep}.
Given an example, if we want to predict which location traveler would like to go at 18:00 pm and we have training time-oriented vector $ (2, 2, 2, 3, 2, 41, 41, 2, 3, 5)$ for 10 hours and from 8:00 am to 18:00 pm. We can take vector $ (2, 2, 2, 3, 2, 41, 41, 2, 3)$ as training set while 5 as targeted label as visitor is at location 5 at 18:00 pm. Therefore we can de-composite the time-oriented Matrix to be training matrix and labour vector

The second step is to use a prediction method to generate a model and then input the test data into model to generate the test label which is the predicted result. Afterwards we can visualize the result using several visualization approaches.The prediction method is the most crucial part. In this paper we use Recurrent Neural Net  \cite{lecun2015deep} to serve as our predictor as its advantage to deal with time series data. Also the classical classification approaches K-Nearest Neighborhood (KNN), Naive Bayes (NB), Decision Tree (DT), Random Forest (RF), Support Vector Machine (SVM) and Adaptive Boosting (AdaBoost) will be used for evaluation.

\section{Experiment}
In this part, we will introduce two of our experiments. For one, it is a mobile user commercial promotion visualization. For another, it is visualization of abnormal pattern of a nature preserve.

\subsection{Visual Analytics for Mobile User Commercial Promotion}
\subsubsection{Data Set}
This data set is from a Chinese Mobile company records the users use the mobile phone stations to visit the IP address in the real time. The data set is formed in $<userId, TimeStamp, Station, positon, IP>$.

\subsubsection{Outstanding User Detection} 
\begin{figure}[!h]	
	\centering
    \includegraphics[width=\linewidth]{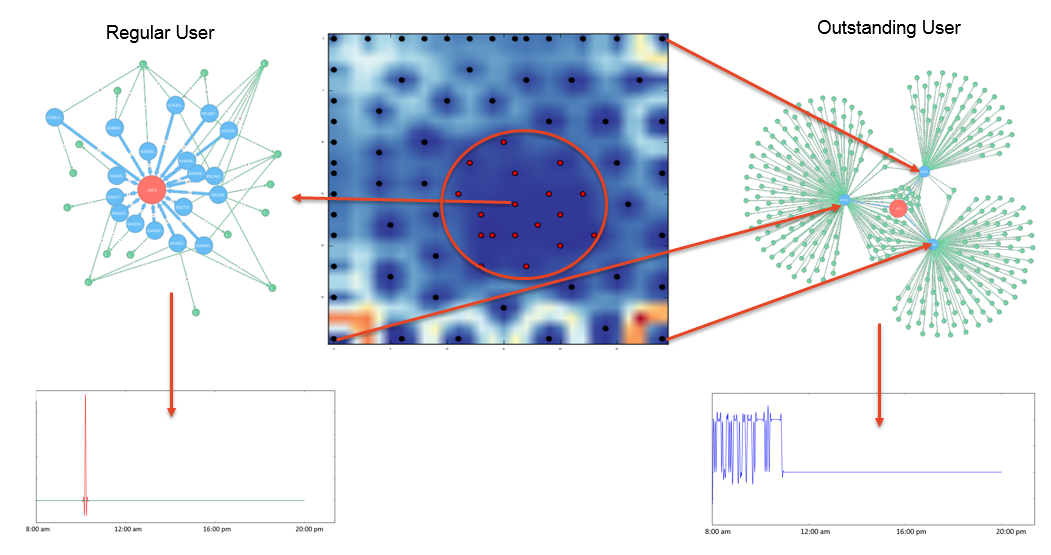}
    \caption{Clustering Result of Mobile user, the left side is the regular user, that user(blue) only use the station(red) visiting IP(green) in a small frequency and the time-spent is short too, while the right side indicates that outstanding user ho spend long time and have check-in with the station.}
	\label{fig:ex1}
\end{figure}
Using SOM to cluster the visiting pattern, we can get the result. There are around 150 stations activated per day. We query one station and cluster the visiting pattern for a week  (5 working days) to view the result. As shown in Figure \ref{fig:ex1}, the down left, down right and up right corners are shown as the outstanding users while the others can be interpreted as regular users.By further visualizing the visiting pattern of both outstanding and regular users, we see that, outstanding users use the station with a large number of frequency and keep activity on the station in quite long time. While the patterns of regular users illustrate that they only use the station with a very small number and activity time is very short too. Therefore, we know that if we want to have a commercial promotion, we should put more efforts on those outstanding users as they are more likely see the advertisements with more time spent on using the mobile station.

\subsubsection{Station Usage Prediction}
\begin{figure}[!h]
\centering
\includegraphics[width=\linewidth]{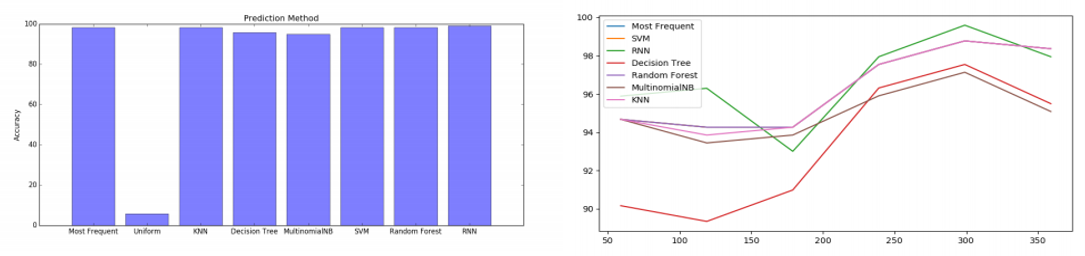}
\caption{\label{fig:eva1} Evaluation Result for predicting the station usage, left side is overall performance, right side is accuracy through time, for both, RNN shows the best performance.}
\end{figure}
\begin{figure}[!h]
\centering
\includegraphics[width=\linewidth]{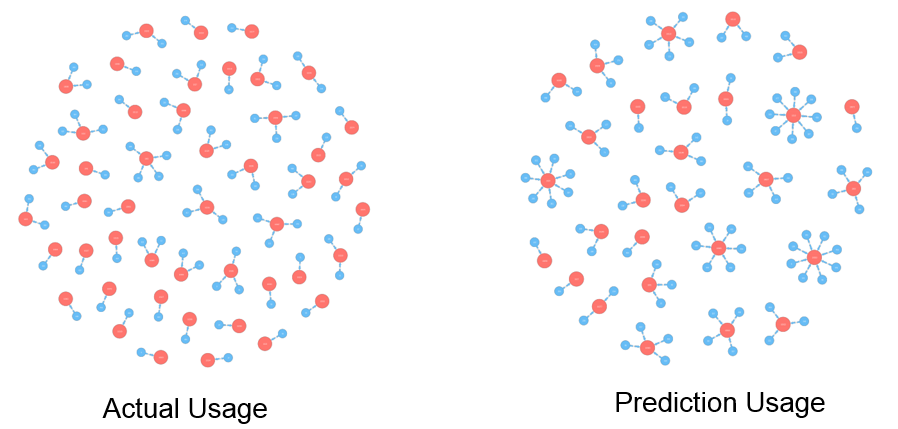}
\caption{\label{fig:pre1} Actual Usage of station (left) and Predicted Usage, indicates the user(blue) checks in station(red) at 17:00 pm.}
\end{figure}
Figure \ref{fig:eva1} shows the prediction accuracy among the different methods. From the result, except the uniform method, other methods achieve very outstanding performance. The highest one is still obtained by RNN with 98\% following by KNN, most frequency, RF and SVM with 97.5\%, 97.2\%, 96.4\% and 96.2\%. It is also interesting to see that, in this data set, most methods can obtain excellent performance. From that, we can also see the accuracy changes through time. RNN (indicated by green line) achieves mostly the highest result among all the time. Especially, when the training set is 300 minutes (5 hours), RNN becomes the best predictor. Also besides RNN, we can see that Decision Tree is also a excellent predictor in this task.
Figure \ref{fig:pre1} shows the actual usage of mobile stations at 17:00 pm, 14 May, 2015 as well as the predicted usage pattern of stations at the same time. We can see that RNN can mostly forecast the usage even with some slight mistakes. In this task, by successfully predicting the usage pattern of stations, we can provide suggestions to the managers. For instance, which station may have a large number of users in next hour, therefore more technical support should be concerned on. For the commercial promotion sides, once we know which station will be busy in next hour, commercial company can put advertisements onto that station.

\subsection{Pattern Analysis for a Nature Preserve}
The data is obtained from VAST Challenge 2017\footnote{\url{http://www.vacommunity.org/VAST+Challenge+2017}} which is trajectory data in a Nature Preserve Boonsong Lekagul. Locations of this preserve can be divided into five categories: Entrances, General-gates, Gates, Ranger-stops, and Camping. The vehicle that traveling around can be divided into 2-axle car  (or motorcycle), 2-axle truck, 3-axle truck, 4-axle  (and above) truck, 2-axle bus and 3-axle bus. Data can be formatted as $<Timestamp,car-id,car-type,gate-name>$.

\begin{figure}[!h]
\centering
\includegraphics[width=\linewidth]{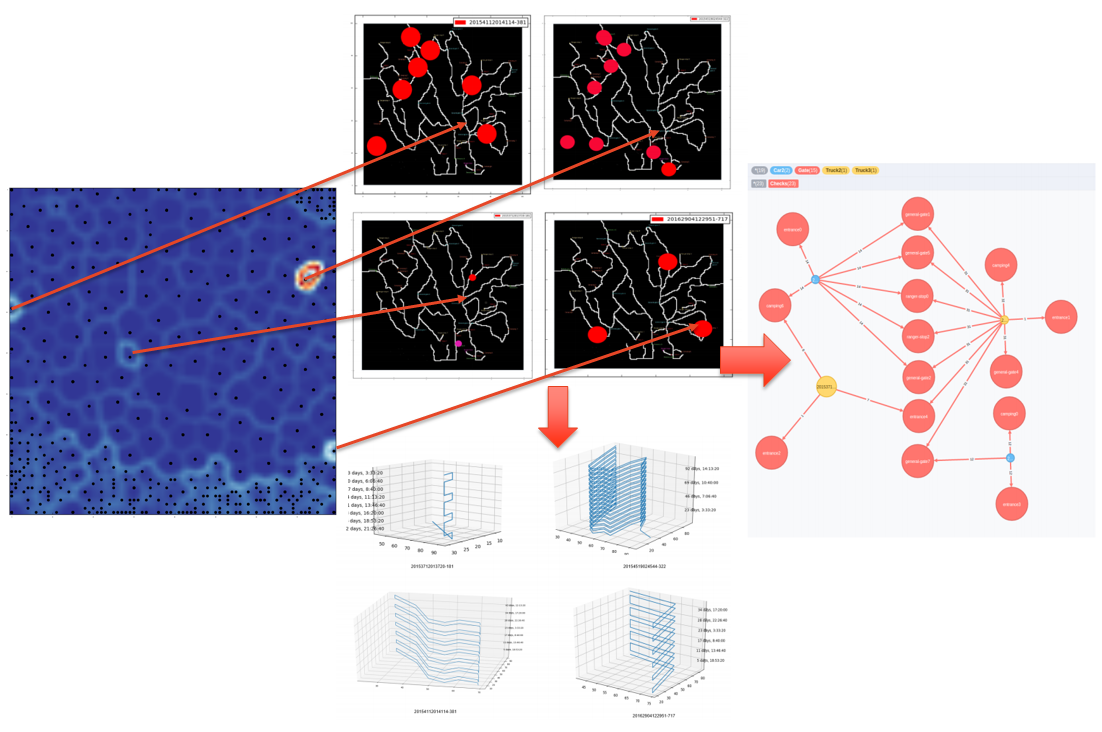}
\caption{\label{fig:ex2} Clustering Result of VAST 2017, there are four outstanding patterns and such patterns visualized by heat map, 3D map and relationship map.}
\end{figure}

\subsubsection{Abnormal Pattern Detection}

Using the SOM clustering. It is quite obvious to see there are four outliers on the U-matrix map shown in Figure \ref{fig:ex2}. These four points are all in the distinct clusters especially right up one which have the largest difference to others.By further visiting the heat map, we can visualize the frequency that those vehicles check-in such relationship can also be indicated by relationship map. 3D time-cube can clearly show trajectories of these four outstanding patterns. We can see that all of these four vehicles travel around in this nature preserve in a long duration. The most outstanding vehicle keeps visiting certain locations for 92 days. We can assume this vehicle will have strongest effect on the wildlife as the longest time it spent in this nature preserve. In other words, as this vehicle occurred over multiple days in the nature preserve, the animals around the path it traveled through would most likely to have been affected for a long time. Such pattern would have much more impact on wildlife than other regular traveling vehicle.

\subsubsection{Traffic Pattern Prediction}
\begin{figure}[!h]
\centering
\includegraphics[width=0.7\linewidth]{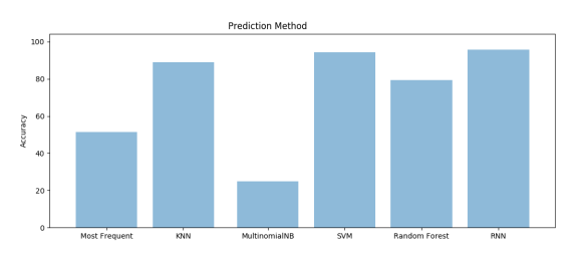}
\caption{\label{fig:eva2} Evaluation prediction result of VAST 2017, RNN shows best performance.}
\end{figure}

\begin{figure}[!h]
\centering
\includegraphics[width=\linewidth]{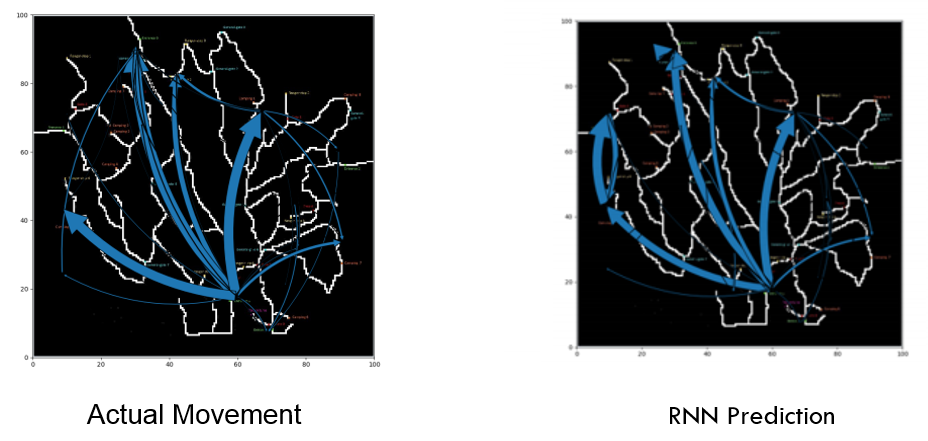}
\caption{\label{fig:pre2} Actual (left) and predicted movement(right) of VAST 2017 19:00- 20:00 pm}
\end{figure}

From Figure \ref{fig:eva2}, we see that RNN get the most outstanding performance with 92.3\%, other three methods SVM, KNN and Random Forest get 87.6\%, 84.2\% and 79\% respectively for the second, third and fourth outstanding performance. Overall, for this data set, the average accuracy for most methods are quite good. Figure \ref{fig:pre2} shows the actual traffic movement from 19:00 pm to 20:00 pm in period from 01 Jan 2016 to 29 May 2016 as well as the RNN predicted movement in the same time period.  In this task, I use the data set from 01 May 2015 to 31 Dec 2015 from 0:00 am to 20:00 pm movement as training data to predict the attendance at 20:00 pm from 01 Jan 2016 to 29 May 2016 . In this task, RNN can successfully outline the four majority trajectories that all start from General Gate 6 but to Camping 5, Ranger Stop 2, General Gate 1 and Camping 1 respectively. As the time is 19:00-20:00 pm, we know that the night time wildlife in those four paths are going get most effects. Correspondingly, we will suggest the manager to take corresponding activities to help the wildlife among these four paths so that mitigate the effects.

\section{Conclusion and Future Work}
The advantage of this framework over conventional methods is that we focus the feature extraction that are clustering and prediction. Such results will indicate essential information to user such as behavior pattern. Moreover, using neural net to render the clustering and predciton rather than simply constructing a visual analytics system, users do not have to conduct expensive experiments and suffer from massive information because of large time-spatial data set. Future work will be concerned on the how to enhance current method and what new methods should be introduced to resolve the temporal-spatial data visual analytics.

% \bibliographystyle{splncs_srt}
% \bibliography{mybibfile}

\end{document}